\pdfoutput=1

\documentclass[11pt]{article}

\usepackage[preprint]{coling}

\usepackage{times}
\usepackage{latexsym}

\usepackage[T1]{fontenc}

\usepackage[utf8]{inputenc}

\usepackage{microtype}

\usepackage{inconsolata}

\usepackage{graphicx}

\usepackage{multirow}
\usepackage{scrextend}

\usepackage{xcolor}

\usepackage{caption}
\title{Enhancing NER Performance in Low-Resource Pakistani Languages using Cross-Lingual Data Augmentation}

\author{Toqeer Ehsan, Thamar Solorio \\
Department of Natural Language Processing \\
MBZUAI\\ 
Abu Dhabi, United Arab Emirates \\
  \texttt{\{toqeer.ehsan, thamar.solorio\}@mbzuai.ac.ae} \\}

\begin{document} 
\maketitle
\begin{abstract}
Named Entity Recognition (NER), a fundamental task in Natural Language Processing (NLP), has shown significant advancements for high-resource languages. However, due to a lack of annotated datasets and limited representation in Pre-trained Language Models (PLMs), it remains understudied and challenging for low-resource languages.  
To address these challenges, we propose a data augmentation technique that generates culturally plausible sentences and experiments on four low-resource Pakistani languages; Urdu, Shahmukhi, Sindhi, and Pashto. By fine-tuning multilingual masked Large Language Models (LLMs), our approach demonstrates significant improvements in NER performance for Shahmukhi and Pashto. We further explore the capability of generative LLMs for NER and data augmentation using few-shot learning.
\end{abstract}

\section{Introduction}
The performance of Named Entity Recognition (NER) in low-resource languages faces challenges due to the scarcity of annotated datasets and insufficient coverage in masked Large Language Models (LLMs) \cite{subedi2024exploring}. 
Causal LLMs, on the other hand, demonstrate their performance by achieving moderate scores for NER \cite{chen2023robust, ye2023comprehensive}. These challenges make it difficult to develop effective NLP applications and highlight the need of focused effort to improve the applicability of these models on available datasets for low-resource languages.

\begin{figure}[t]
\centering
\includegraphics[width=\columnwidth]{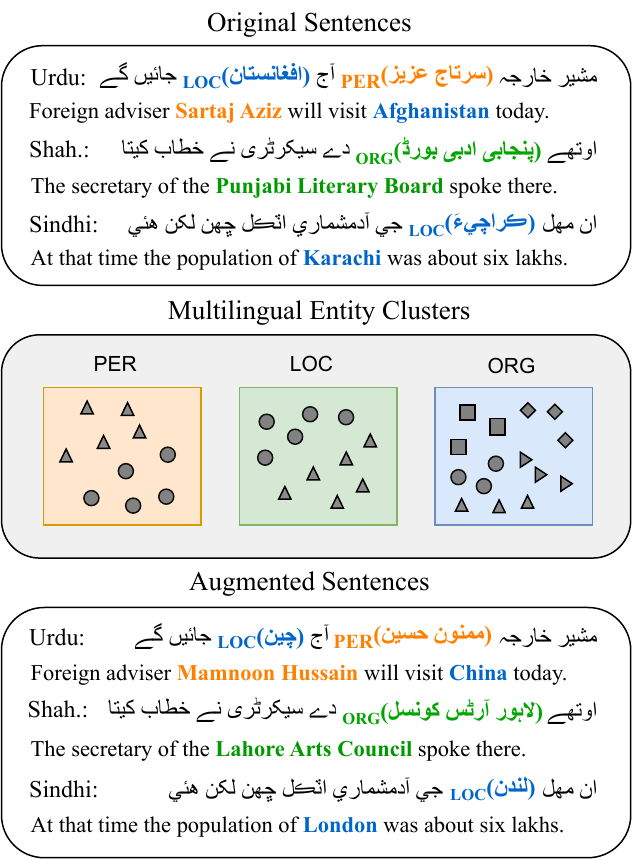}
\caption{\textcolor{black}{Examples of clustering-based data augmentation applied to three sample sentences. Entity mentions are represented in orange, blue and green colors.}
} 
  \label{fig:1}
\end{figure}




Data augmentation approaches could be effective to enhance the NER datasets for low-resource languages.
One such approach is the Easy Data Augmentation (EDA) \cite{wei2019eda}, that offers simple and effective techniques, including synonym replacement, random insertion, random swap, and random deletion \cite{khalid2023using, liu2023improving, litake2024inditext}. However, EDA can produce linguistically implausible text lacking verbal agreement based on gender and number.
Additionally, EDA may produce out-of-context or offensive data for culturally sensitive content. 
This can affect the generalizability and learning of NER models.
We aim to enhance NER performance for Pakistani low-resource languages by employing effective data augmentation as shown in Figure~\ref{fig:1}. 

\begin{figure}[ht]
  \includegraphics[width=\columnwidth]{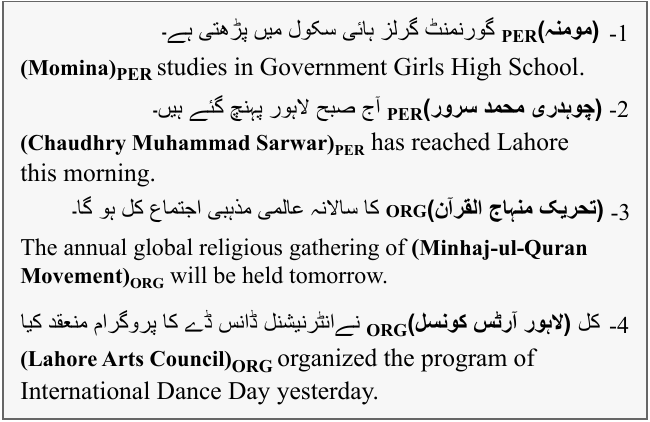}
  \caption{Sample Urdu sentences for the analysis of EDA. Named entities are highlighted in bold.}
  \label{fig:2}
\end{figure}

Four Urdu sentences are shown in Figure~\ref{fig:2}, illustrating the problem of implausibility. Urdu, Shahmukhi and Sindhi require verbal agreements, and augmenting entities from the sentences 1 and 2, could result in disagreements. 
\textit{Momina} (Nom.Fem.Sg) is a feminine name that has agreement with the verb \textit{paRHtI} (study.Hab.Fem.Sg), while \textit{Chaudhry Muhammad Sarwar} (Nom.Masc.Sg) is a masculine name that agrees with the verb \textit{gaE} (go.Past.Masc.Sg.Hon). Replacing these named entities can produce implausible text; for instance, the sentence \textit{Chaudhry Muhammad Sarwar studies in Government Girls High School} would violate the verbal agreement rules of the language. The named entities in the last two sentences are considered opposites within the community, and replacing such named entities can produce text that is very offensive to the native community. The generated sentences remain grammatically correct but create contextual ambiguity.


We propose a cross-lingual data augmentation technique by clustering named entities as shown in Figure~\ref{fig:1}. This technique improves the quality of culturally sensitive content and grammar of the augmented text. We performed unsupervised entity clustering and entity replacement by aligning clusters for the source and candidate named entities of each type. NER experiments were conducted for low-resource settings as well as for entire datasets. We compared the results with EDA-based and generative augmentation methods for mono- and multilingual settings by fine-tuning the Glot500 \cite{imani2023glot500} and XLM-RoBERTa \cite{DBLP:journals/corr/abs-1911-02116} models. Shahmukhi and Pashto datasets demonstrated significant improvements, producing F\textsubscript{1} scores of 88.06 and 88.29 with increases of 5.53 and 1.81 points, respectively.

Zero- or few-shot learning is relevant in low-resource scenarios where even augmented datasets are limited in size.
We explore the capabilities of causal LLMs to perform NER and data augmentation for our low-resource languages using few-shot learning. The key contributions of this paper are as follows:

\begin{itemize}
\setlength\itemsep{0em}

\item We propose a novel cross-lingual augmentation technique that uses cluster dictionaries to produce culturally and linguistically plausible augmentations.
\item We demonstrate the effectiveness of the proposed technique in multilingual NER experiments by utilizing cross-lingual representations.
\item We provide insights into the potential of causal LLMs to perform NER and data augmentation for low-resource languages using few-shot learning.
\end{itemize}


\begin{figure*}[t]
\centering
\includegraphics[width=\linewidth]{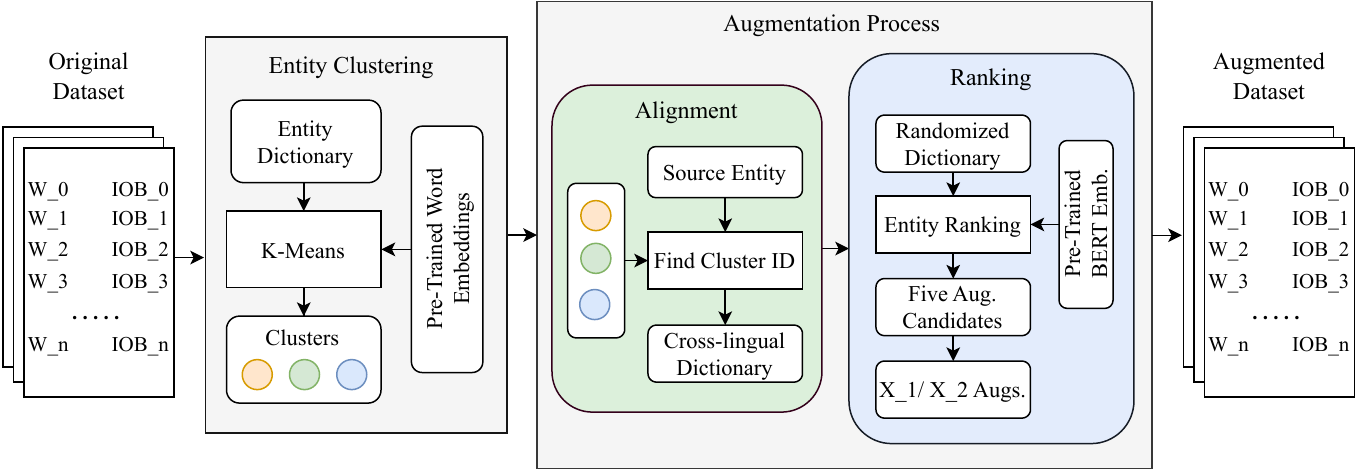}
\caption {Cluster-based data augmentation process, which contains three phases. The entity clustering phase extracts unsupervised clusters for each entity type, alignment phase aligns cluster dictionaries with respect to the source (original) entities and the final phase ranks the source entity mentions with the best candidate. The original dataset corresponds to the manually annotated dataset, while the augmented dataset is the updated version obtained through the augmentation process.}
  \label{fig:3}
\end{figure*}

\section{Related Work}
\label{sec_2}
Manually annotated corpora are crucial for achieving state-of-the-art results in NER \cite{mayhew2023universal}. Cross-lingual transfer also supports generalization and enhances the performance of models \cite{ding2024improving, mo2024mcl, cotterell2024low, le2024constrained, pmlr-v119-hu20b}. Data augmentation techniques enhance the size and learning capabilities of datasets for low-resource languages \cite{litake2024inditext, ye2024llm, lancheros2024data}. For the task of NER, three data augmentation methods are mainly used; Easy Data Augmentation (EDA) \cite{wei2019eda} and its variants, translation-based methods and generative LLMs. EDA-based techniques demonstrate enhanced NER performance for low-resource languages \cite{litake2024inditext}. The data augmentation quality can be enhanced by using contextualized word embeddings \cite{torres2024experimental} and cosine similarity \cite{bartolini2022cosiner}.

Data augmentation based on back-translation has shown improvements for code-switched NER \cite{sabty2021data}. The translation-based data augmentation technique that performs cross-lingual entity augmentation also improves the performance of NER models \cite{liu2021mulda, lancheros2024data, chen2022frustratingly}. 

The capabilities of causal LLMs are being explored for data augmentation \cite{evuru2024coda, ye2024llm} and underlying NLP tasks such as NER \cite{naguib2024few, villena2024llmner, lu2024large}. Generative data augmentation techniques have demonstrated improvements \cite{evuru2024coda, liu2022low, ye2024llm}. Masking-based generative methods have produced better NER results by generating more plausible data augmentations \cite{song2024ropda}.

Causal LLMs are further employed to perform NER with zero- and few-shot learning \cite{naguib2024few, villena2024llmner} as an alternative approach to data augmentation. These models are also progressing in various text domains \cite{lu2024large, monajatipoor2024llms}. These advancements highlight the need to investigate the capabilities of these models for low-resource languages. 


\section{Cross-Lingual Data Augmentation}
\label{sec_4}
The languages selected in this work are topologically related and culturally similar. In terms of named entities, they share similar names, locations and organizations. Given these similarities, cross-lingual representation could be helpful in improving the performance of NER for the regional languages. Additionally, data augmentation techniques have shown improvements for low-resource languages, but EDA-based methods are blunt and may produce culturally offensive and/or ungrammatical sentences by replacing entities with the other entities of the same type without any additional semantic information. \textcolor{black}{Out of 100 randomly selected sentences, 32 instances of verbal disagreements and three of sensitive religious named entities were found. These percentages estimate the occurrence of such issues in the augmented data.}
To address these issues, we propose a data augmentation technique that generates more sensible sentences and produces competitive NER performance for the selected low-resource languages. 
The next section describes our proposed technique, followed by descriptions of EDA-based random replacement and generative approaches.



\subsection{Cluster-based\textsubscript{Aug.}}
\label{sec:3-1}
We propose a hybrid data augmentation technique inspired by EDA, combined with the application of unsupervised entity clustering. The technique consists of three phases; entity clustering, alignment, and ranking as illustrated by  Figure~\ref{fig:3}.

\paragraph{\textit{Entity Clustering}}
Named entities were clustered using context-free word embeddings from pre-trained models \cite{grave2018learning, tehseen2023shahmukhi}, where each word has a single embedding regardless of its context which are helpful in clustering process. We employed the K-Means clustering algorithm to cluster entities based on their embeddings and cosine similarity. While clustering is an unsupervised method, we interpreted these clustering representing specific categories for each entity type. To evaluate the effectiveness of the approach, we manually assessed the unsupervised clustering of 200 entities for each entity type in Urdu. The person and location types were categorized into two clusters; masculine and feminine for persons, and country/continent and city/places for locations. In contrast, named entities from the organization type were grouped into ten clusters; entertainment, financial, health/education, justice/govt, news, politics, religious, water/electricity, abbreviations and miscellaneous. The accuracies for correctly clustered named entities were 86.0\% for persons, 87.5\% for locations, and 84.5\% for organizations, as determined through manual evaluation. 
The K-means clustering approach was implemented using NLTK's \textit{KMeansClusterer} to categorize named entity embeddings into distinct groups. The clustering process utilized cosine distance as the similarity metric, ensuring that entities with similar vector representations were grouped together effectively. To enhance the stability and robustness of the clustering process, we performed 25 repetitions. For clustering, separate dictionaries of unique named entities were created based on the splits of annotated training sets.

We achieve a single feature vector by averaging the vectors for each token in an entity of the location and organization types.
However, person names have a specific pattern in Pakistani culture. 
The first name usually belongs to the individual, followed by a family name.
A feminine first name is typically followed by a masculine name, that could be the name of the father, tribe, caste, or creed. For instance, in the entity mention \textit{Madiha Khalid}, \textit{Madiha} is the feminine name followed by the masculine name \textit{Khalid}. 
Similarly, many names, particularly masculine names, begin with a title representing a designation, tribe, caste, or creed. We prepared a list of these titles to filter them out and used first names to obtain feature vectors. This approach improved the performance of clustering. 


\paragraph{\textit{Alignment}}
The prepared clusters are aligned between the source and candidate entity mentions. The source entity refers to the original entity mention in the dataset, while the candidate entity is the one selected to replace the source entity. In the alignment phase, the cluster ID of the source entity is determined by looking it up in the manually identified clusters. A dictionary containing unique named entities from the corresponding cluster is then passed to the next phase.

\paragraph{\textit{Ranking}}
The ranking procedure is performed in two steps. In the first step, an entity is selected from a randomized cluster dictionary by computing highest cosine similarity with respect to the source entity mention. Unlike the clustering process, contextualized word embeddings from Glot500-base, which has data coverage of all our selected languages, are used to select similar candidate entities. This step generates five augmented sentences for each original sentence. In the second step, micro F\textsubscript{1} score is computed for augmented sentences to assess their plausibility, using Glot500-base model fine-tuned on multilingual datasets. This pretrained model automatically validates each generated candidate. The tokens of each augmented sentence are fed into the model to predict the named entities. The sentence with the highest F\textsubscript{1} score is selected to be a part of the augmented dataset. The F\textsubscript{1} score is computed by treating the model output as the predicted annotation, while the manually annotated named entities in the augmented sentence serve as a reference in the process. We further prepared multiple augmented datasets by including one sentence with the highest score (X\textsubscript{1}), two sentences with top two scores (X\textsubscript{2}) and all augmented sentences with an F\textsubscript{1} score of 1.0. 

\subsection{Random Replacement (EDA-RR\textsubscript{Aug.})}
\label{sec:3-2}
The random replacement data augmentation is a straightforward approach which is based on EDA methods \cite{wei2019eda}. The augmentation process has two steps; 1) take all sentences in the training data with labeled named entities, 2) for each entity mention in a sentence, replace it with a named entity of the same type. 
The second step continues until all entity mentions in a sentence are replaced randomly. As a result, a new augmented dataset is produced, which is added to the training set to enhance its size and diversity. This method is simple and efficient to implement, but it may produce contextually implausible text that could be incorrect or offensive to the community.

\subsection{Generative\textsubscript{Aug.}} 
\label{sec:3-3}
To add the contextual information in the data augmentation, we performed generative data augmentation using LLaMA3 \cite{touvron2023llama} with few-shot learning. The approach is similar to the entity-level augmentation proposed by \citet{ye2024llm}. We employed instruction-finetuned version of LLaMA3 (LLaMA3-8B-Instruct). We selected LLaMA3 due to its open-access nature and strong few-shot learning capabilities. LLaMA3 has been trained on a diverse multilingual corpus, but its direct exposure to Pakistani languages is limited. However, Urdu is a widely spoken language with significant online resources, LLaMA3 demonstrates moderate performance in generating Urdu text.
We constructed a prompt by providing three examples containing each entity type and instructed the model to replace entity mentions with similar entities. The augmentation was performed for low-resource training sets due to time and resource constraints. The prompt that we used for data augmentation is given below:\\

\begin{addmargin}[1em]{1em}
\textit{You are an expert in augmenting data for named entities for Urdu language. The input contains the ORIGINAL TEXT followed by the AUGMENTED TEXT. Perform augmentation by replacing named entities with new entities of the same type and return the AUGMENTED TEXT. Three examples are given for your reference:\\
EXAMPLE 1: \\
{ORIGINAL TEXT}: \\
{AUGMENTED TEXT}:}
\end{addmargin}

\section{Languages and Datasets}
\label{sec_3}
Pakistan is home to many widely spoken languages, each with unique linguistic characteristics and cultural significance. Urdu is the national language of Pakistan that has 232 million speakers worldwide. Shahmukhi (Punjabi), Sindhi, and Pashto have 67, 30 and 40 million speakers, respectively \cite{eberhard2024ethnologue}. 
These languages pose several challenges for the task of NER, such as absence of capitalization, contextual ambiguity, flexible word-order, and agglutinating nature \cite{khalid2023using, ehsan2021development, ahmed2024enriching}.
The statistics of the selected datasets are shown in Table~\ref{tab:datasets}. 
\textcolor{black}{Despite the larger sample sizes in Shahmukhi, Sindhi, and Urdu datasets, they face limited domain coverage, incomplete NER labels, low sentence-to-entity ratio, and noisy annotations, underscoring their low-resource status.} \textcolor{black}{The MK-PUCIT, Shahmukhi and SiNER datasets were released without validation sets; therefore, we used 10\% of the train sets for validation.}

\paragraph{Urdu:}
Being in the \textit{Vital} category \cite{eberhard2024ethnologue}, Urdu is relatively resource-rich compared to the regional languages. Several NER datasets are available for Urdu with different data annotations and sizes \cite{khana2016named, hussain2008resources, jahangir2012n, malik2017urdu}. However, we experimented with Urdu-Wikiann \cite{rahimi-etal-2019-massively,lovenia2024seacrowd} and MK-PUCIT \cite{kanwal2019urdu}, which are larger datasets annotated with coarse-grained named entities; person, location and organization.
 
\begin{table}[t]
\centering
\small
\begin{tabular}{cllll}
\hline
\textbf{Lang./Dataset} & \textbf{Type} & \textbf{Train} & \textbf{Test} & \textbf{Val.} \\
\hline
\multirow{4}{*}{} & PER & 6,839 & 363 & 340 \\
Urdu / & LOC & 6,891 & 334 & 352 \\
Urdu-Wikiann & LOC & 6,891 & 334 & 352 \\
 & ORG & 6,759 & 323 & 327 \\
 & \# Sents. & 20,000 & 1,000 & 1,000 \\
\hline
\multirow{4}{*}{} & PER & 11,965 & 5,215 & – \\
Urdu / & LOC & 23,880 & 8,380 & – \\
MK-PUCIT & ORG & 8,665 & 3,014 & – \\
 & \# Sents. & 24,080 & 16,609 & – \\
\hline
\multirow{4}{*}{} & PER & 4,655 & 1,957 & – \\
Punjabi / & LOC & 1,855 & 648 & – \\
Shahmukhi & ORG & 538 & 236 & – \\
 & \# Sents. & 13,547 & 5,821 & – \\
\hline
\multirow{4}{*}{} & PER & 12,894 & 5,564 & – \\
Sindhi / & LOC & 2,769 & 630 & – \\
SiNER & ORG & 1,331 & 891 & – \\
 & \# Sents. & 31,870 & 7,418 & – \\
\hline
\multirow{4}{*}{} & PER & 32 & 28 & 39 \\
Pashto / & LOC & 37 & 45 & 45 \\
Pashto-Wikiann & ORG & 43 & 38 & 33 \\
 & \# Sents. & 100 & 100 & 100\\
\hline
\end{tabular}
\caption{Type-wise statistics of the datasets for Urdu, Shahmukhi, Sindhi and Pashto.}
\label{tab:datasets}
\end{table}

\paragraph{Shahmukhi:}
There is only one NER dataset available for Shahmukhi, which has been annotated using person, location, and organization types \cite{ahmad2020named}. The quality of the dataset was further enhanced by using the BIO annotation scheme \cite{tehseen2023shahmukhi}. The dataset contained some annotation inconsistencies. To ensure the validity of our NER results, we manually reviewed and corrected the annotations in one thousand sentences from the test set. While this review process was conducted to enhance the reliability of our evaluation.

\paragraph{Sindhi:}
\citet{ali2020siner} released the first large annotated dataset for the Sindhi language called SiNER. 
We experimented with three coarse-grained entity types to make it compatible with the other datasets. 

\paragraph{Pashto:}
Pashto lacks in fundamental language processing tools \cite{eberhard2024ethnologue}. 
We used the Pashto dataset from Wikiann \cite{rahimi-etal-2019-massively} that contains 100 sentences for train, test and validation sets. Since the dataset was automatically annotated and exhibited some annotation inconsistencies, we reviewed the test set manually to ensure valid NER results.

\section{Experimental Setup}
\label{sec_5}
We conducted NER experiments designed to improve performance for low-resource languages, where supervised models often struggle due to limited annotated datasets. This research addresses three key questions; 1) How effective are data augmentation techniques to enhance NER for low-resource languages? 2) Do cross-lingual data representations improve NER performance in multilingual settings?
3) How does few-shot learning compare to fully supervised models as an alternative to data augmentation?
We hypothesize that cross-lingual representations, combined with multilingual datasets improve NER results for topologically related and culturally similar languages.


\subsection{NER Models and Architectures}
For our NER experiments, we employed two pre-trained multilingual masked language models: Glot500-base \cite{imani2023glot500} and XLM-RoBERTa-large \cite{DBLP:journals/corr/abs-1911-02116}.
\begin{itemize}
\setlength\itemsep{0em}
\item Glot500-base supports over 500 languages and is based on RoBERTa's \cite{DBLP:journals/corr/abs-1911-02116} architecture. It uses transformer-based contextualized token embeddings and is particularly designed for low-resource languages like Urdu, Shahmukhi, Sindhi, and Pashto.

\item XLM-RoBERTa-large is another transformer-based multilingual models that supports 100 languages, including Urdu, Sindhi, and Pashto. It is pre-trained on massive multilingual text corpora using masked language modeling (MLM) objectives.
\end{itemize}

To fine-tune these models for NER, we added a token classification layer on the top of the final transformer layer which receives the hidden states from the last layer of the model and computes the multi-class probability distribution over the entity classes for each token. This setup classifies tokens into person, location and organization categories. 

We fine-tuned both models on mono- and multilingual datasets to investigate their performance for NER for low-resource setting by including 100, 200, 500 and 1000 train samples. Additionally, we experimented with the data augmentation techniques to further improve NER performance for low-resource languages. 

\subsection{Few-Shot Learning with Causal Models}
While the primary focus of this paper is on data augmentation techniques to enhance NER performance in low-resource languages, we also explore few-shot learning as an alternative approach. Although various causal LLMs have recently been evaluated for the task of NER, they still struggle to compete with state-of-the-art supervised models \cite{naguib2024few, villena2024llmner, lu2024large}. This raises a research question; how well do these models perform in low-resource languages? 



We performed NER by using a few-shot learning approach by prompting LLaMA3-8B-Instruct\footnote{https://huggingface.co/meta-llama/Meta-Llama-3-8B-Instruct} and Mistral-7B-Instruct-v0.3\footnote{https://huggingface.co/mistralai/Mistral-7B-v0.3} which are instruction tuned. LLaMA3-8B is trained on 15 trillion tokens with a context length of 8K. Mistral-7B also has the same context length but its training size is not disclosed. We created a prompt, similar to Generative\textsubscript{Aug.}, describing details of the task by providing three examples for each language (Appendix~\ref{sec:appendixE}). The inputs and outputs were formatted as sequences of texts and NER labels. For erroneous outputs, the number of labels matching the number of tokens in the input was selected for evaluation.
We evaluated the performance of both causal models on 1,000 sentences from each dataset. 

\section{Results and Discussion}
\label{sec_6}

\begin{table*}[t]
\small
\centering
\begin{tabular}{llcccccccc}
\hline
\multicolumn{2}{c}{\textbf{Monolingual Settings}} & \multicolumn{4}{c}{\textbf{Glot500-base}} & \multicolumn{4}{c}{\textbf{XLM-RoBERTa-large}} \\
\multicolumn{1}{l}{\textbf{Dataset}} & \textbf{Augmentation} & \textbf{100} & \textbf{200} & \textbf{500} & \textbf{1000} & \textbf{100} & \textbf{200} & \textbf{500} & \textbf{1000} \\
\hline
\multirow{4}{*}{Urdu-Wikiann} & Original dataset & 70.93 & 77.23 & 83.51 & 80.24 & 72.77 & 71.21 & 84.21 & \textbf{87.21} \\
 & Generative\textsubscript{Aug.} & \textbf{77.13} & {79.29} & 84.24 & \textbf{86.81} & \textbf{79.66} & \textbf{83.85} & \textbf{85.01} & 85.50 \\
 & EDA-RR\textsubscript{Aug.} & 74.87 & 77.42 & \textbf{84.57} & 85.87 & 71.75 & 80.27 & 82.79 & 85.84 \\
 & Cluster-based\textsubscript{Aug.} & 76.62 & \textbf{81.00} & 83.78 & 85.31 & 75.24 & 80.79 & 84.30 & {85.97} \\
 
\hline

\multirow{4}{*}{Shahmukhi} & Original dataset & 59.62 & 65.27 & 71.92 & 75.44 & 53.67 & 59.44 & 70.65 & 75.58 \\
 & Generative\textsubscript{Aug.} & 53.83 & 62.45 & 69.85 & 74.89 & 58.81 & 58.64 & 66.96 & 74.68 \\
 & EDA-RR\textsubscript{Aug.} & 58.44 & 63.98 & 70.34 & 73.87 & 51.95 & 64.75 & 72.40 & 75.32 \\
 & Cluster-based\textsubscript{Aug.} & \textbf{60.78} & \textbf{68.03} & \textbf{73.17} & \textbf{77.11} & \textbf{59.61} & \textbf{65.89} & \textbf{74.19} & \textbf{77.40} \\

\hline

\multirow{4}{*}{SiNER} & Original dataset & 62.25 & 69.61 & 75.82 & \textbf{80.27} & 73.63 & \textbf{78.16} & 81.22 & 82.80 \\
 & Generative\textsubscript{Aug.} & 53.76 & 60.64 & 69.09 & 73.76 & 64.12 & 71.58 & 73.81 & 77.09 \\
 & EDA-RR\textsubscript{Aug.} & 64.69 & \textbf{72.29} & 73.50 & 72.65 & \textbf{75.40} & 75.22 & 80.01 & 83.00 \\
 & Cluster-based\textsubscript{Aug.} & \textbf{65.64} & 71.17 & \textbf{76.88} & 79.46 & {74.27} & 75.96 & \textbf{81.60} & \textbf{84.48} \\

\hline
\multirow{4}{*}{Pashto-Wikiann} & Original dataset & 32.86 &--&--&--& 44.24 &--&--&--\\
& Generative\textsubscript{Aug.} & 45.66 & -- &--&--& 45.26 &--&--&--\\
& EDA-RR\textsubscript{Aug.} & 45.75 &--&--&--& 48.92 &--&--&--\\
& Cluster-based\textsubscript{Aug.} & \textbf{48.54} &--&--&--& \textbf{50.00} &--&--&--\\
\hline
\multicolumn{2}{c}{\textbf{Multilingual Settings}} & \multicolumn{4}{c}
{\textbf{Glot500-base}} & \multicolumn{4}{c}{\textbf{XLM-RoBERTa-large}} \\
\multicolumn{1}{l}{\textbf{Dataset}} & \textbf{Augmentation} & \textbf{100} & \textbf{200} & \textbf{500} & \textbf{1000} & \textbf{100} & \textbf{200} & \textbf{500} & \textbf{1000} \\
\hline
\multirow{4}{*}{Urdu-Wikiann} & Original dataset & 73.12 & 74.82 & \textbf{84.45} & 84.90 & 63.32 & 78.25 & \textbf{82.33} & 80.97 \\
 & Generative\textsubscript{Aug.} & \textbf{78.92} & 79.67 & 84.16 & \textbf{85.77} & 77.78 & \textbf{80.93} & 82.07 & \textbf{85.43} \\
 & EDA-RR\textsubscript{Aug.} & 72.75 & 77.43 & 83.04 & 83.98 & \textbf{79.13} & 78.70 & 81.10 & 82.02 \\
 & Cluster-based\textsubscript{Aug.} & 76.83 & \textbf{82.25} & {84.35} & 85.34 & 78.60 & 79.49 & 81.14 & 83.21 \\

\hline
\multirow{4}{*}{Shahmukhi} & Original dataset & 65.33 & 69.21 & 75.84 & 79.38 & 56.87 & 67.85 & 72.27 & 76.66 \\
 & Generative\textsubscript{Aug.} & 64.32 & 68.45 & 74.46 & 77.45 & 65.69 & 69.23 & 74.68 & 78.21 \\
 & EDA-RR\textsubscript{Aug.} & 66.23 & 69.34 & 74.48 & 78.32 & 65.90 & 69.44 & \textbf{75.86} & 77.26 \\
 & Cluster-based\textsubscript{Aug.} & \textbf{68.88} & \textbf{73.47} & \textbf{77.51} & \textbf{80.01} & \textbf{67.83} & \textbf{71.38} & 73.79 & \textbf{78.90} \\

\hline
\multirow{4}{*}{SiNER} & Original dataset & 62.35 & 67.78 & 73.85 & 78.83 & 67.37 & 74.02 & 76.42 & 79.23 \\
 & Generative\textsubscript{Aug.} & 58.53 & 65.30 & 71.78 & 73.59 & 56.76 & 68.78 & 75.19 & 76.96 \\
 & EDA-RR\textsubscript{Aug.} & 64.72 & 69.71 & 74.30 & 77.84 & 69.79 & 74.48 & 76.06 & 79.77 \\
 & Cluster-based\textsubscript{Aug.} & \textbf{66.76} & \textbf{73.22} & \textbf{76.26} & \textbf{79.61} & \textbf{71.99} & \textbf{75.24} & \textbf{78.40} & \textbf{80.45} \\

\hline
\multirow{4}{*}{Pashto-Wikiann} & Original dataset & 62.26 & 67.68 & 73.68 & 78.58 & 67.01 & 73.79 & 76.22 & 78.96 \\
 & Generative\textsubscript{Aug.} & 58.51 & 65.19 & 71.62 & 73.43 & 65.68 & 68.66 & 74.98 & 76.73 \\
 & EDA-RR\textsubscript{Aug.} & 64.66 & 69.53 & 74.12 & 77.59 & 69.60 & 74.32 & 75.86 & 79.53 \\
 & Cluster-based\textsubscript{Aug.} & \textbf{66.63} & \textbf{73.12} & \textbf{76.05} & \textbf{79.35} & \textbf{71.78} & \textbf{74.98} & \textbf{78.17} & \textbf{80.21}\\
\hline
\end{tabular}
\caption{Micro-F\textsubscript{1} scores of fine-tuned multilingual Glot500-base and XLM-RoBERTa-large models for NER in low-resource settings. The results of the cluster-based augmentation are compared against the original training set, generative augmentation from LLaMa3 (Generative\textsubscript{Aug.}) and EDA - Random Replacement (EDA-RR\textsubscript{Aug.}). 
}
\label{tab:lowresresults}
\end{table*}

\begin{table*}[t]
\centering
\small
\begin{tabular}{llcccccc}

\hline
\multicolumn{2}{c}{\textbf{Monolingual Settings}} & \multicolumn{3}{c}{\textbf{Glot500-base}} & \multicolumn{3}{c}{\textbf{XLM-RoBERTa-large}} \\

\multicolumn{1}{l}{\textbf{Dataset}} & \textbf{Augmentation} & \multicolumn{1}{l}{\textbf{Precision}} & \multicolumn{1}{l}{\textbf{Recall}} & \multicolumn{1}{l}{\textbf{F\textsubscript{1} Score}} & \multicolumn{1}{l}{\textbf{Precision}} & \multicolumn{1}{l}{\textbf{Recall}} & \multicolumn{1}{l}{\textbf{F\textsubscript{1} Score}} \\
\hline
\multirow{3}{*}{Urdu-Wikiann} & Original dataset & 95.46 & 95.86 & \textbf{95.66} & 95.80 & 96.75 & \textbf{96.28} \\
 & EDA-RR\textsubscript{Aug.} & 94.89 & 96.38 & 95.63 & 96.08 & 96.08 & 96.08 \\
 & Cluster-based\textsubscript{Aug.} & 94.51 & 94.61 & 94.56 & 93.97 & 94.34 & 94.16 \\

\hline
\multirow{3}{*}{Shahmukhi} & Original dataset & 79.12 & 73.92 & 76.44 & 80.77 & 73.40 & 76.91 \\
 & EDA-RR\textsubscript{Aug.} & 85.58 & 76.96 & 81.04 & 85.55 & 79.76 & \textbf{82.55} \\
 & Cluster-based\textsubscript{Aug.} & 84.04 & 82.23 & \textbf{83.13} & 86.52 & 78.71 & 82.43 \\

\hline
\multirow{3}{*}{SiNER} & Original dataset & 90.50 & 85.69 & 88.03 & 88.78 & 89.12 & 88.95 \\
 & EDA-RR\textsubscript{Aug.} & 88.66 & 87.88 & \textbf{88.27} & 88.14 & 90.10 & \textbf{89.11} \\
 & Cluster-based\textsubscript{Aug.} & 87.49 & 88.82 & 88.15 & 87.50 & 89.68 & 88.58 \\
\hline
\multirow{3}{*}{Pashto-Wikiann} & Original dataset & 51.55 & 32.29 & 39.71 & 49.74 & 38.86 & 43.63 \\
 & EDA-RR\textsubscript{Aug.} & 46.45 & 47.77 & \textbf{47.10} & 48.19 & 53.45 & \textbf{50.68} \\
 & Cluster-based\textsubscript{Aug.} & 43.93 & 46.96 & 45.40 & 54.23 & 46.54 & 50.09 \\
 \hline
\multicolumn{2}{c}{\textbf{Multilingual Settings}} & \multicolumn{3}{c}{\textbf{Glot500-base}} & \multicolumn{3}{c}{\textbf{XLM-RoBERTa-large}} \\
\multicolumn{1}{l}{\textbf{Dataset}} & \textbf{Augmentation} & \multicolumn{1}{l}{\textbf{Precision}} & \multicolumn{1}{l}{\textbf{Recall}} & \multicolumn{1}{l}{\textbf{F\textsubscript{1} Score}} & \multicolumn{1}{l}{\textbf{Precision}} & \multicolumn{1}{l}{\textbf{Recall}} & \multicolumn{1}{l}{\textbf{F\textsubscript{1} Score}} \\
\hline
\multirow{3}{*}{Urdu-Wikiann} & Original dataset & 95.93 & 96.38 & 96.16 & 96.09 & 96.43 & \textbf{96.26} \\
& EDA-RR\textsubscript{Aug.} & 96.10 & 96.75 & \textbf{96.42} & 95.02 & 95.58 & 95.30 \\
& Cluster-based\textsubscript{Aug.} & 96.07 & 96.18 & 96.12 & 96.23 & 96.28 & 96.25 \\
\hline
\multirow{3}{*}{Shahmukhi} & Original dataset & 83.63 & 80.70 & 82.14 & 83.36 & 81.71 & 82.53 \\
 & EDA-RR\textsubscript{Aug.} & 87.98 & 83.43 & 85.64 & 88.51 & 83.62 & 86.00 \\
 & Cluster-based\textsubscript{Aug.} & 89.03 & 85.50 & \textbf{87.22} & 89.29 & 86.85 & \textbf{88.06} \\
\hline
\multirow{3}{*}{SiNER} & Original dataset & 87.99 & 84.82 & 86.37 & 87.12 & 86.35 & 86.73 \\
 & EDA-RR\textsubscript{Aug.} & 88.01 & 86.91 & 87.46 & 90.52 & 86.33 & 88.38 \\
 & Cluster-based\textsubscript{Aug.} & 89.19 & 86.69 & \textbf{87.92} & 89.33 & 87.80 & \textbf{88.56} \\
\hline
\multirow{3}{*}{Pashto-Wikiann} & Original dataset & 87.78 & 84.63 & 86.18 & 86.77 & 86.18 & 86.48 \\
& EDA-RR\textsubscript{Aug.} & 87.51 & 86.72 & 87.12 & 90.15 & 85.94 & 88.00 \\
& Cluster-based\textsubscript{Aug.} & 89.00 & 86.29 & \textbf{87.62} & 89.14 & 87.45 & \textbf{88.29} \\
\hline
\end{tabular}
\caption{Micro-F\textsubscript{1} scores of fine-tuned multilingual Glot500-base and XLM-RoBERTa-large models for complete datasets. The results of the cluster-based augmentation are compared against the original training sets and EDA - Random Replacement (EDA-RR\textsubscript{Aug.}). Improved scores are highlighted in bold.}
\label{tab:completeresults}
\end{table*}

We use micro F-scores to ensure a balanced evaluation of NER performance across all entity types. Table~\ref{tab:lowresresults} presents Micro-F\textsubscript{1} score for low-resource NER experiments using monolingual and multilingual data settings. The training sets contain 100, 200, 500 and 1,000 samples for each dataset. In the multilingual settings, we combined training samples from all selected languages (Urdu, Shahmukhi, Sindhi, and Pashto). To maintain balanced representation, we ensured that each language contributed an equal number of samples in low-resource scenarios.
The results are presented from fine-tuned Glot500-base and XLM-RoBERTa-large models. 
Similarly, Table~\ref{tab:completeresults} shows NER results for the entire datasets. \textcolor{black}{The training samples in all augmented datasets were doubled in one iteration, and the NER results are presented after this iteration. Further analysis from multiple iterations is presented in the Appendix~\ref{sec:appendixB}.} 

Our data augmentation technique improved NER results for low-resource languages by reducing the generation of grammatically implausible and culturally offensive content. The augmentation technique helps maintain semantics and cultural appropriateness, that highly impacted the model performance. The model trained on the augmented datasets demonstrated higher generalizability due to less exposure to the contextually implausible information. This confirms that grammatically and contextually inappropriate data can degrade the model performance by introducing noise and reducing its ability to generalize effectively. The following paragraphs present a comparison of data augmentation techniques for each dataset. 

\paragraph{\textit{Urdu-Wikiann}}

The Urdu-Wikiann dataset demonstrates inconsistent performance for different augmentation techniques, which is caused by three main reasons. First, Urdu is a resource-rich language compared to the other three regional languages and fine-tuning using cross-lingual data augmentation enhances its diversity, but does not significantly impact NER results due to the large size of the dataset.
Second, causal LLMs, such as LLaMA3 have better support for Urdu compared to the other three languages as Urdu dataset shows improvements using Generative\textsubscript{Aug.} method.
Third, the Urdu-Wikiann dataset is an automatically annotated dataset that may have some inconsistencies \cite{mayhew2023universal} which can limit the effectiveness of cross-lingual augmentation.

\paragraph{\textit{Shahmukhi}}
The Shahmukhi dataset demonstrates consistent performance with cluster-based data augmentation as the proposed method generates plausible augmentations that leads to improved results. The fine-tuned XLM model produced a state-of-the-art F\textsubscript{1} score of 88.06 in multilingual settings using the BIO annotation scheme, which outperforms the previous best score of 75.55 \cite{tehseen2023shahmukhi}. 

However, Generative\textsubscript{Aug.} decreased NER performance for Shahmukhi. The causal model produced various augmentations that violated entity types, resulting in incorrect labeling. The low scores indicate that multilingual causal LLMs have limited support for low-resource languages. 
The cluster-based data augmentation technique outperformed other two augmentation methods in both monolingual and multilingual experiments.


\paragraph{\textit{SiNER}}
For the Sindhi dataset, the cluster-based cross-lingual augmentation improved NER results in a multilingual setting by utilizing cross-lingual representations. This approach introduced linguistic variation and diversity that enhanced the models' ability to generalize. For the entire dataset, EDA-RR\textsubscript{Aug.} demonstrated improved results by adding cross-lingual entities that enriched the training set, making it a suitable augmentation technique for Sindhi in a monolingual training setup. However, Generative\textsubscript{Aug.} had a negative impact on all low-resource training sets, highlighting limited capabilities of causal LLMs for low-resource languages. Sindhi's use of Arabic script with additional unique letters, unlike Urdu, Shahmukhi, and Pashto, may negatively impact multilingual fine-tuning




\paragraph{\textit{Pashto-Wikiann}}
The Pashto-Wikiann dataset demonstrates significant improvements with data augmentation techniques, especially in a multilingual setup, except for Generative\textsubscript{Aug.}. 
The best reported F\textsubscript{1} score for Pashto is 82.0 achieved from an HMM-based tagger \cite{momand2020comparative}. 
By using cluster-based augmentation, the multilingual fine-tuned Glot500 and XLM models produced F\textsubscript{1} scores of 87.62 and 88.29, respectively. 
However, these findings should be interpreted with caution due to the small size of the training and evaluation sets, which may limit the generalizability of the results.

\paragraph{\textit{Few-Shot Learning}}
Table~\ref{tab:fewshotresults} presents NER results obtained from causal LLMs using few-shot learning. The performance of both LLaMA-3-8B and Mistral-7B on low-resource languages is not remarkable. LLaMa-3 performed better for Shahmukhi; however, its performance on Urdu, a relatively high-resource language, is quite low. The few-shot NER results indicate that causal LLMs are still far behind in NER for low-resource languages. 

\begin{table}[ht]
\small
\centering
\begin{tabular}{lccc}
\hline
\multicolumn{4}{c}{\textbf{LLaMA-3-8B-Instruct}} \\
\textbf{Dataset} & \textbf{Precision} & \textbf{Recall} & \textbf{F\textsubscript{1} Score} \\
\hline
Urdu-Wikiann & 20.13 & 24.26 & 22.00 \\
Shahmukhi & 74.63 & 72.06 & 73.32 \\
SiNER & 39.98 & 48.66 & 43.89 \\
Pashto-Wikiann & 48.46 & 56.76 & 52.28 \\
\hline
\multicolumn{4}{c}{\textbf{Mistral-7B-Instruct-v0.3}} \\
\textbf{Dataset} & \textbf{Precision} & \textbf{Recall} & \textbf{F\textsubscript{1} Score}\\
\hline
Urdu-Wikiann & 42.54 & 45.29 & 43.87 \\
Shahmukhi & 41.49 & 47.13 & 44.13 \\
SiNER & 27.02 & 38.40 & 31.72 \\
Pashto-Wikiann & 47.29 & 54.95 & 50.83 \\
\hline
\end{tabular}
\caption{Micro-F\textsubscript{1} scores by few-shot learning NER from LLaMA3-8B-Insruct and Mistral-7B-Instruct-v0.3. Both models have been evaluated for 1,000 sentences from each dataset except Pashto-Wikiann that has only 100 samples.}
\label{tab:fewshotresults}
\end{table}

\subsection{Limitations}
Despite demonstrating significant advantages in the application of cross-lingual data augmentation, this study has a few limitations. The Shahmukhi, SiNER and MK-PUCIT datasets contain some annotation inconsistencies and errors that affect the overall performance of the models. Furthermore, the cluster-based data augmentation technique used entity clusters by employing an unsupervised clustering algorithm. The accuracy of the clustering process poses a limitation on the quality of the augmentation. 
Future work should focus on improving the annotation quality and consistency of such datasets. 

\section{Conclusion}
\label{sec_7}
This study explored various data augmentation techniques and their effect on the task of NER for low-resource languages. We used pre-trained LLMs on mono- and multilingual setups. Our findings highlight that cluster-based data augmentation improves NER performance for Shahmukhi, Sindhi and Pashto datasets by incorporating linguistically plausible text and cross-lingual diversity. Urdu-Wikiann, an automatically annotated dataset, does not take advantage of cross-lingual augmentations. Generative augmentation shows improved results on Urdu, while have a negative impact on the other three regional languages. Few-shot learning with causal models reveal their current limitations for low-resource languages when used for data augmentation and NER. Overall, the research emphasizes the potential of hybrid data augmentation techniques to enhance NER performance for low-resource languages. 



\bibliography{coling_latex}

\begin{thebibliography}{45}
\providecommand{\natexlab}[1]{#1}

\bibitem[{Ahmad et~al.(2020)Ahmad, Malik, Shahzad, Aslam, Iqbal, Nawaz, and Bukhari}]{ahmad2020named}
Muhammad~Tayyab Ahmad, Muhammad~Kamran Malik, Khurram Shahzad, Faisal Aslam, Asif Iqbal, Zubair Nawaz, and Faisal Bukhari. 2020.
\newblock {N}amed {E}ntity {R}ecognition and {C}lassification for {P}unjabi {S}hahmukhi.
\newblock \emph{ACM Transactions on Asian and Low-Resource Language Information Processing (TALLIP)}, 19(4):1--13.

\bibitem[{Ahmed et~al.(2024)Ahmed, Huang, Arafat, and Hameed}]{ahmed2024enriching}
Anil Ahmed, Degen Huang, Syed~Yasser Arafat, and Imran Hameed. 2024.
\newblock Enriching {U}rdu {N}er with {BERT} {E}mbedding, {D}ata {A}ugmentation, and {H}ybrid {E}ncoder-{CNN} {A}rchitecture.
\newblock \emph{ACM Transactions on Asian and Low-Resource Language Information Processing}, 23(4):1--38.

\bibitem[{Ali et~al.(2020)Ali, Lu, and Xu}]{ali2020siner}
Wazir Ali, Junyu Lu, and Zenglin Xu. 2020.
\newblock {S}i{NER}: {A} {L}arge {D}ataset for {S}indhi {N}amed {E}ntity {R}ecognition.
\newblock In \emph{Proceedings of the Twelfth Language Resources and Evaluation Conference}, pages 2953--2961.

\bibitem[{Bartolini et~al.(2022)Bartolini, Moscato, Postiglione, Sperl{\`\i}, and Vignali}]{bartolini2022cosiner}
Ilaria Bartolini, Vincenzo Moscato, Marco Postiglione, Giancarlo Sperl{\`\i}, and Andrea Vignali. 2022.
\newblock {COSINER}: {CO}ntext {SI}milarity data augmentation for {N}amed {E}ntity {R}ecognition.
\newblock In \emph{International Conference on Similarity Search and Applications}, pages 11--24. Springer.

\bibitem[{Chen et~al.(2023)Chen, Ye, Zu, Xu, Zheng, Peng, Zhou, Gui, Zhang, and Huang}]{chen2023robust}
Xuanting Chen, Junjie Ye, Can Zu, Nuo Xu, Rui Zheng, Minlong Peng, Jie Zhou, Tao Gui, Qi~Zhang, and Xuanjing Huang. 2023.
\newblock How {R}obust is {GPT}-3.5 to {P}redecessors? {A} {C}omprehensive {S}tudy on {L}anguage {U}nderstanding {T}asks.
\newblock \emph{arXiv preprint arXiv:2303.00293}.

\bibitem[{Chen et~al.(2022)Chen, Jiang, Ritter, and Xu}]{chen2022frustratingly}
Yang Chen, Chao Jiang, Alan Ritter, and Wei Xu. 2022.
\newblock {F}rustratingly {E}asy {L}abel {P}rojection for {C}ross-lingual {T}ransfer.
\newblock \emph{arXiv preprint arXiv:2211.15613}.

\bibitem[{Conneau et~al.(2019)Conneau, Khandelwal, Goyal, Chaudhary, Wenzek, Guzm{\'{a}}n, Grave, Ott, Zettlemoyer, and Stoyanov}]{DBLP:journals/corr/abs-1911-02116}
Alexis Conneau, Kartikay Khandelwal, Naman Goyal, Vishrav Chaudhary, Guillaume Wenzek, Francisco Guzm{\'{a}}n, Edouard Grave, Myle Ott, Luke Zettlemoyer, and Veselin Stoyanov. 2019.
\newblock \href {https://arxiv.org/abs/1911.02116} {{U}nsupervised {C}ross-lingual {R}epresentation {L}earning at {S}cale}.
\newblock \emph{CoRR}, abs/1911.02116.

\bibitem[{Cotterell and Duh(2024)}]{cotterell2024low}
Ryan Cotterell and Kevin Duh. 2024.
\newblock {L}ow-{R}esource {N}amed {E}ntity {R}ecognition with {C}ross-{L}ingual, {C}haracter-level {N}eural {C}onditional {R}andom {F}ields.
\newblock \emph{arXiv preprint arXiv:2404.09383}.

\bibitem[{Ding et~al.(2024)Ding, Wei, Qu, and Chen}]{ding2024improving}
Zhuojun Ding, Wei Wei, Xiaoye Qu, and Dangyang Chen. 2024.
\newblock Improving {P}seudo {L}abels with {G}lobal-{L}ocal {D}enoising {F}ramework for {C}ross-lingual {N}amed {E}ntity {R}ecognition.
\newblock \emph{arXiv preprint arXiv:2406.01213}.

\bibitem[{Eberhard and Fennig(2024)}]{eberhard2024ethnologue}
Gary F.~Simons Eberhard, David~M. and Charles~D. Fennig. 2024.
\newblock \href {http://www.ethnologue.com} {Ethnologue: \uppercase{L}anguages of the world}.
\newblock \emph{SIL International}, 27.

\bibitem[{Ehsan and Hussain(2021)}]{ehsan2021development}
Toqeer Ehsan and Sarmad Hussain. 2021.
\newblock Development and {E}valuation of an {U}rdu {T}reebank ({CLE-UTB}) and a {S}tatistical {P}arser.
\newblock \emph{Language Resources and Evaluation}, 55(2):287--326.

\bibitem[{Evuru et~al.(2024)Evuru, Ghosh, Kumar, Tyagi, Manocha et~al.}]{evuru2024coda}
Chandra Kiran~Reddy Evuru, Sreyan Ghosh, Sonal Kumar, Utkarsh Tyagi, Dinesh Manocha, et~al. 2024.
\newblock {C}o{D}a: {C}onstrained {G}eneration based {D}ata {A}ugmentation for {L}ow-{R}esource {NLP}.
\newblock \emph{arXiv preprint arXiv:2404.00415}.

\bibitem[{Grave et~al.(2018)Grave, Bojanowski, Gupta, Joulin, and Mikolov}]{grave2018learning}
Edouard Grave, Piotr Bojanowski, Prakhar Gupta, Armand Joulin, and Tomas Mikolov. 2018.
\newblock {L}earning {W}ord {V}ectors for 157 {L}anguages.
\newblock In \emph{Proceedings of the International Conference on Language Resources and Evaluation (LREC 2018)}.

\bibitem[{Hu et~al.(2020)Hu, Ruder, Siddhant, Neubig, Firat, and Johnson}]{pmlr-v119-hu20b}
Junjie Hu, Sebastian Ruder, Aditya Siddhant, Graham Neubig, Orhan Firat, and Melvin Johnson. 2020.
\newblock \href {https://proceedings.mlr.press/v119/hu20b.html} {{XTREME}: A massively multilingual multi-task benchmark for evaluating cross-lingual generalisation}.
\newblock In \emph{Proceedings of the 37th International Conference on Machine Learning}, volume 119 of \emph{Proceedings of Machine Learning Research}, pages 4411--4421. PMLR.

\bibitem[{Hussain(2008)}]{hussain2008resources}
Sarmad Hussain. 2008.
\newblock Resources for {U}rdu language processing.
\newblock In \emph{Proceedings of the 6th workshop on Asian Language Resources}.

\bibitem[{Imani et~al.(2023)Imani, Lin, Kargaran, Severini, Sabet, Kassner, Ma, Schmid, Martins, Yvon et~al.}]{imani2023glot500}
Ayyoob Imani, Peiqin Lin, Amir~Hossein Kargaran, Silvia Severini, Masoud~Jalili Sabet, Nora Kassner, Chunlan Ma, Helmut Schmid, Andr{\'e}~FT Martins, Fran{\c{c}}ois Yvon, et~al. 2023.
\newblock Glot500: {S}caling {M}ultilingual {C}orpora and {L}anguage {M}odels to 500 {L}anguages.
\newblock \emph{arXiv preprint arXiv:2305.12182}.

\bibitem[{Jahangir et~al.(2012)Jahangir, Anwar, Bajwa, and Wang}]{jahangir2012n}
Faryal Jahangir, Waqas Anwar, Usama~Ijaz Bajwa, and Xuan Wang. 2012.
\newblock N-gram and {G}azetteer {L}ist based {N}amed {E}ntity {R}ecognition for {U}rdu: {A} {S}carce {R}esourced {L}anguage.
\newblock In \emph{Proceedings of the 10th Workshop on Asian Language Resources}, pages 95--104.

\bibitem[{Kanwal et~al.(2019)Kanwal, Malik, Shahzad, Aslam, and Nawaz}]{kanwal2019urdu}
Safia Kanwal, Kamran Malik, Khurram Shahzad, Faisal Aslam, and Zubair Nawaz. 2019.
\newblock Urdu {N}amed {E}ntity {R}ecognition: {C}orpus {G}eneration and {D}eep {L}earning {A}pplications.
\newblock \emph{ACM Transactions on Asian and Low-Resource Language Information Processing (TALLIP)}, 19(1):1--13.

\bibitem[{Khalid et~al.(2023)Khalid, Murtaza, and Abbas}]{khalid2023using}
Hamza Khalid, Ghulam Murtaza, and Qaiser Abbas. 2023.
\newblock {U}sing {D}ata {A}ugmentation and {B}idirectional {E}ncoder {R}epresentations from {T}ransformers for {I}mproving {P}unjabi {N}amed {E}ntity {R}ecognition.
\newblock \emph{ACM Transactions on Asian and Low-Resource Language Information Processing}, 22(6):1--13.

\bibitem[{Khana et~al.(2016)Khana, Daudb, Nasira, and Amjada}]{khana2016named}
Wahab Khana, Ali Daudb, Jamal~A Nasira, and Tehmina Amjada. 2016.
\newblock Named {E}ntity {D}ataset for {U}rdu {N}amed {E}ntity {R}ecognition {T}ask.
\newblock \emph{Language \& Technology}, 51.

\bibitem[{Lancheros et~al.(2024)Lancheros, Corpas~Pastor, and Mitkov}]{lancheros2024data}
Brayan~Stiven Lancheros, Gloria Corpas~Pastor, and Ruslan Mitkov. 2024.
\newblock {D}ata {A}ugmentation and {T}ransfer {L}earning for {C}ross-lingual {N}amed {E}ntity {R}ecognition in the {B}iomedical {D}omain.
\newblock \emph{Language Resources and Evaluation}, pages 1--20.

\bibitem[{Le et~al.(2024)Le, Chen, Ritter, and Xu}]{le2024constrained}
Duong~Minh Le, Yang Chen, Alan Ritter, and Wei Xu. 2024.
\newblock {C}onstrained {D}ecoding for {C}ross-lingual {L}abel {P}rojection.
\newblock \emph{arXiv preprint arXiv:2402.03131}.

\bibitem[{Litake et~al.(2024)Litake, Yagnik, and Labhsetwar}]{litake2024inditext}
Onkar Litake, Niraj Yagnik, and Shreyas Labhsetwar. 2024.
\newblock {I}ndi{T}ext {B}oost: {T}ext {A}ugmentation for {L}ow {R}esource {I}ndia {L}anguages.
\newblock \emph{arXiv preprint arXiv:2401.13085}.

\bibitem[{Liu et~al.(2022)Liu, Chen, and Xu}]{liu2022low}
Jian Liu, Yufeng Chen, and Jinan Xu. 2022.
\newblock Low-{R}esource {NER} by {D}ata {A}ugmentation {W}ith {P}rompting.
\newblock In \emph{IJCAI}, pages 4252--4258.

\bibitem[{Liu et~al.(2021)Liu, Ding, Bing, Joty, Si, and Miao}]{liu2021mulda}
Linlin Liu, Bosheng Ding, Lidong Bing, Shafiq Joty, Luo Si, and Chunyan Miao. 2021.
\newblock {M}ul{DA}: {A} {M}ultilingual {D}ata {A}ugmentation {F}ramework for {L}ow-{R}esource {C}ross-{L}ingual {NER}.
\newblock In \emph{Proceedings of the 59th Annual Meeting of the Association for Computational Linguistics and the 11th International Joint Conference on Natural Language Processing (Volume 1: Long Papers)}, pages 5834--5846.

\bibitem[{Liu and Cui(2023)}]{liu2023improving}
Wenzhong Liu and Xiaohui Cui. 2023.
\newblock Improving {N}amed {E}ntity {R}ecognition for {S}ocial {M}edia with {D}ata {A}ugmentation.
\newblock \emph{Applied Sciences}, 13(9):5360.

\bibitem[{Lovenia et~al.(2024)Lovenia, Mahendra, Akbar, Miranda, Santoso, Aco, Fadhilah, Mansurov, Imperial, Kampman, Moniz, and Others.}]{lovenia2024seacrowd}
Holy Lovenia, Rahmad Mahendra, Salsabil~Maulana Akbar, Lester James~V. Miranda, Jennifer Santoso, Elyanah Aco, Akhdan Fadhilah, Jonibek Mansurov, Joseph~Marvin Imperial, Onno~P. Kampman, Joel Ruben~Antony Moniz, and Others. 2024.
\newblock \href {https://arxiv.org/abs/2406.10118} {{SEAC}rowd: {A} {M}ultilingual {M}ultimodal {D}ata {H}ub and {B}enchmark {S}uite for {S}outheast {A}sian {L}anguages}.
\newblock \emph{arXiv preprint arXiv: 2406.10118}.

\bibitem[{Lu et~al.(2024)Lu, Li, Wen, Wang, Wang, and Liu}]{lu2024large}
Qiuhao Lu, Rui Li, Andrew Wen, Jinlian Wang, Liwei Wang, and Hongfang Liu. 2024.
\newblock {L}arge {L}anguage {M}odels {S}truggle in {T}oken-{L}evel {C}linical {N}amed {E}ntity {R}ecognition.
\newblock \emph{arXiv preprint arXiv:2407.00731}.

\bibitem[{Malik(2017)}]{malik2017urdu}
Muhammad~Kamran Malik. 2017.
\newblock {U}rdu {N}amed {E}ntity {R}ecognition and {C}lassification {S}ystem using {A}rtificial {N}eural {N}etwork.
\newblock \emph{ACM Transactions on Asian and Low-Resource Language Information Processing (TALLIP)}, 17(1):1--13.

\bibitem[{Mayhew et~al.(2023)Mayhew, Blevins, Liu, {\v{S}}uppa, Gonen, Imperial, Karlsson, Lin, Ljube{\v{s}}i{\'c}, Miranda et~al.}]{mayhew2023universal}
Stephen Mayhew, Terra Blevins, Shuheng Liu, Marek {\v{S}}uppa, Hila Gonen, Joseph~Marvin Imperial, B{\"o}rje~F Karlsson, Peiqin Lin, Nikola Ljube{\v{s}}i{\'c}, Lester~James Miranda, et~al. 2023.
\newblock {U}niversal {NER}: {A} {G}old-{S}tandard {M}ultilingual {N}amed {E}ntity {R}ecognition {B}enchmark.
\newblock \emph{arXiv preprint arXiv:2311.09122}.

\bibitem[{Mo et~al.(2024)Mo, Yang, Liu, Wang, Chen, Wang, and Li}]{mo2024mcl}
Ying Mo, Jian Yang, Jiahao Liu, Qifan Wang, Ruoyu Chen, Jingang Wang, and Zhoujun Li. 2024.
\newblock {MCL-NER}: {C}ross-{L}ingual {N}amed {E}ntity {R}ecognition via {M}ulti-{V}iew {C}ontrastive {L}earning.
\newblock In \emph{Proceedings of the AAAI Conference on Artificial Intelligence}, volume~38, pages 18789--18797.

\bibitem[{Momand et~al.(2020)Momand, Waseeb, and Rai}]{momand2020comparative}
Rafiullah Momand, Shakirullah Waseeb, and Ahmad Masood~Latif Rai. 2020.
\newblock A {C}omparative {S}tudy of {D}ictionary-based and {M}achine {L}earning-based {N}amed {E}ntity {R}ecognition in {P}ashto.
\newblock In \emph{Proceedings of the 4th International Conference on Natural Language Processing and Information Retrieval}, pages 96--101.

\bibitem[{Monajatipoor et~al.(2024)Monajatipoor, Yang, Stremmel, Emami, Mohaghegh, Rouhsedaghat, and Chang}]{monajatipoor2024llms}
Masoud Monajatipoor, Jiaxin Yang, Joel Stremmel, Melika Emami, Fazlolah Mohaghegh, Mozhdeh Rouhsedaghat, and Kai-Wei Chang. 2024.
\newblock {LLM}s in {B}iomedicine: {A} {S}tudy on {C}linical {N}amed {E}ntity {R}ecognition.
\newblock \emph{arXiv preprint arXiv:2404.07376}.

\bibitem[{Naguib et~al.(2024)Naguib, Tannier, and N{\'e}v{\'e}ol}]{naguib2024few}
Marco Naguib, Xavier Tannier, and Aur{\'e}lie N{\'e}v{\'e}ol. 2024.
\newblock Few {S}hot {C}linical {E}ntity {R}ecognition in {T}hree {L}anguages: {M}asked {L}anguage {M}odels {O}utperform {LLM} {P}rompting.
\newblock \emph{arXiv preprint arXiv:2402.12801}.

\bibitem[{Rahimi et~al.(2019)Rahimi, Li, and Cohn}]{rahimi-etal-2019-massively}
Afshin Rahimi, Yuan Li, and Trevor Cohn. 2019.
\newblock \href {https://www.aclweb.org/anthology/P19-1015} {Massively {M}ultilingual {T}ransfer for {NER}}.
\newblock In \emph{Proceedings of the 57th Annual Meeting of the Association for Computational Linguistics}, pages 151--164, Florence, Italy. Association for Computational Linguistics.

\bibitem[{Sabty et~al.(2021)Sabty, Omar, Wasfalla, Islam, and Abdennadher}]{sabty2021data}
Caroline Sabty, Islam Omar, Fady Wasfalla, Mohamed Islam, and Slim Abdennadher. 2021.
\newblock {D}ata {A}ugmentation {T}echniques on {A}rabic {D}ata for {N}amed {E}ntity {R}ecognition.
\newblock \emph{Procedia Computer Science}, 189:292--299.

\bibitem[{Song et~al.(2024)Song, Shen, and Zhao}]{song2024ropda}
Sihan Song, Furao Shen, and Jian Zhao. 2024.
\newblock {R}o{PDA}: {R}obust {P}rompt-{B}ased {D}ata {A}ugmentation for {L}ow-{R}esource {N}amed {E}ntity {R}ecognition.
\newblock In \emph{Proceedings of the AAAI Conference on Artificial Intelligence}, volume~38, pages 19017--19025.

\bibitem[{Subedi et~al.(2024)Subedi, Regmi, Bal, and Acharya}]{subedi2024exploring}
Bipesh Subedi, Sunil Regmi, Bal~Krishna Bal, and Praveen Acharya. 2024.
\newblock {E}xploring the {P}otential of {L}arge {L}anguage {M}odels ({LLM}s) for {L}ow-resource {L}anguages: {A} study on {N}amed-{E}ntity {R}ecognition ({NER}) and {P}art-{O}f-{S}peech ({POS}) {T}agging for {N}epali {L}anguage.
\newblock In \emph{Proceedings of the 2024 Joint International Conference on Computational Linguistics, Language Resources and Evaluation (LREC-COLING 2024)}, pages 6974--6979.

\bibitem[{Tehseen et~al.(2023)Tehseen, Ehsan, Liaqat, Kong, Ali, and Al-Fuqaha}]{tehseen2023shahmukhi}
Amina Tehseen, Toqeer Ehsan, Hannan~Bin Liaqat, Xiangjie Kong, Amjad Ali, and Ala Al-Fuqaha. 2023.
\newblock {S}hahmukhi {N}amed {E}ntity {R}ecognition by using {C}ontextualized {W}ord {E}mbeddings.
\newblock \emph{Expert Systems with Applications}, 229:120489.

\bibitem[{Torres et~al.(2024)Torres, de~Moura, da~Silva, Nascimento, and Mesquita}]{torres2024experimental}
Arthur~Elwing Torres, Edleno~Silva de~Moura, Altigran~Soares da~Silva, Mario~A Nascimento, and Filipe Mesquita. 2024.
\newblock An {E}xperimental {S}tudy on {D}ata {A}ugmentation {T}echniques for {N}amed {E}ntity {R}ecognition on {L}ow-{R}esource {D}omains.

\bibitem[{Touvron et~al.(2023)Touvron, Lavril, Izacard, Martinet, Lachaux, Lacroix, Rozi{\`e}re, Goyal, Hambro, Azhar et~al.}]{touvron2023llama}
Hugo Touvron, Thibaut Lavril, Gautier Izacard, Xavier Martinet, Marie-Anne Lachaux, Timoth{\'e}e Lacroix, Baptiste Rozi{\`e}re, Naman Goyal, Eric Hambro, Faisal Azhar, et~al. 2023.
\newblock {LL}a{MA}: {O}pen and {E}fficient {F}oundation {L}anguage {M}odels.
\newblock \emph{arXiv preprint arXiv:2302.13971}.

\bibitem[{Villena et~al.(2024)Villena, Miranda, and Aracena}]{villena2024llmner}
Fabi{\'a}n Villena, Luis Miranda, and Claudio Aracena. 2024.
\newblock llm{NER}:({Z}ero| {F}ew)-{S}hot {N}amed {E}ntity {R}ecognition, {E}xploiting the {P}ower of {L}arge {L}anguage {M}odels.
\newblock \emph{arXiv preprint arXiv:2406.04528}.

\bibitem[{Wei and Zou(2019)}]{wei2019eda}
Jason Wei and Kai Zou. 2019.
\newblock {EDA}: {E}asy {D}ata {A}ugmentation {T}echniques for {B}oosting {P}erformance on {T}ext {C}lassification {T}asks.
\newblock \emph{arXiv preprint arXiv:1901.11196}.

\bibitem[{Ye et~al.(2023)Ye, Chen, Xu, Zu, Shao, Liu, Cui, Zhou, Gong, Shen et~al.}]{ye2023comprehensive}
Junjie Ye, Xuanting Chen, Nuo Xu, Can Zu, Zekai Shao, Shichun Liu, Yuhan Cui, Zeyang Zhou, Chao Gong, Yang Shen, et~al. 2023.
\newblock A {C}omprehensive {C}apability {A}nalysis of {GPT}-3 and {GPT}-3.5 {S}eries {M}odels.
\newblock \emph{arXiv preprint arXiv:2303.10420}.

\bibitem[{Ye et~al.(2024)Ye, Xu, Wang, Zhou, Zhang, Gui, and Huang}]{ye2024llm}
Junjie Ye, Nuo Xu, Yikun Wang, Jie Zhou, Qi~Zhang, Tao Gui, and Xuanjing Huang. 2024.
\newblock {LLM-DAA}: {D}ata {A}ugmentation via {L}arge {L}anguage {M}odels for {F}ew-{S}hot {N}amed {E}ntity {R}ecognition.
\newblock \emph{arXiv preprint arXiv:2402.14568}.

\end{thebibliography}

\newpage

\appendix

\section{MK-PUCIT Dataset}
\label{sec:appendixA}
The MK-PUCIT dataset was released with IO (Inside-Outside) annotation that has some annotation inconsistencies and errors. We converted it to the BIO (Begin-Inside-Outside) scheme automatically. For missing annotations, we extracted dictionaries with unique entities for each entity type from the training set and mapped the missing annotations throughout the dataset. After the mapping process, there was an overall increase of 19.9\% in entity mentions for the train set and an increase of 13.8\% for the test set. This highlights a significant number of missing annotations. Table~\ref{tab:datasets} presents the updated statistics of the MK-PUCIT dataset. 

We performed NER experiments by fine-tuning the Glot500 model and compared the results with different versions of the dataset in mono- and multilingual settings. Table~\ref{tab:mk-pucitresults} shows NER results for the MK-PUCIT. The original dataset, after conversion from IO to BIO scheme, performs with a micro F\textsubscript{1} score of 68.47. By performing the entity mapping for missing annotations, its performance was enhanced by 8.69 points, which is a significant improvement. Its performance remains in the same range in a multilingual setup. F\textsubscript{1} scores for the other three languages are lower compared to Urdu-Wikiann, therefore, we selected the Urdu-Wikiann dataset for multilingual NER experiments in this study.

\begin{table}[ht]
\small
\centering
\begin{tabular}{lccc}
\hline
\multicolumn{4}{c}{\textbf{Monolingual NER}} \\
\textbf{Dataset} & \textbf{Precision} & \textbf{Recall} & \textbf{F\textsubscript{1} Score}\\
\hline
MK-PUCIT\textsubscript{Original} & 74.27 & 63.51 & 68.47 \\
MK-PUCIT\textsubscript{Mapped} & 81.14 & 73.56 & 77.16 \\
\hline
MK-PUCIT\textsubscript{Combined} & 83.26 & 72.27 & 77.37 \\
Shahmukhi & 81.89 & 74.75 & 78.15 \\
SiNER & 81.44 & 79.76 & 80.59 \\
Pashto-Wikiann & 81.32 & 79.62 & 80.46 \\
\hline
\end{tabular}
\caption{NER results by fine-tuning Glot500-base on the MK-PUCIT dataset. The fine-tuned model has been trained on; 1) original dataset after conversion from IO scheme to BIO, 2) with entity mapping for missing annotations, 3) multilingual setup by combining datasets of four languages. }
\label{tab:mk-pucitresults}
\end{table}

\section{Dataset Analysis}
\label{sec:appendixC}
To investigate the capability of pre-trained models to generalize cross-lingual entity representations, we analyzed the ratio of named entities which are common in both training and test sets. The main objective of this analysis is to determine whether the models are only memorizing seen examples or if they are improving generalization in multilingual training setup?. Table~\ref{tab:dataset-analysis} shows type-wise presence of entity mentions from the test sets in the training sets. The analysis is given for both, mono- and multilingual datasets. All four datasets demonstrate a minor increase in seen examples from monolingual to multilingual datasets. The small increase in the ratio of seen entities is evident that the models enhance their learning by generalization and produce better NER results in multilingual setups.

\begin{table*}[t]
\small
\centering
\begin{tabular}{lllll}
\hline
\multicolumn{5}{c}{\textbf{Monolingual Datasets}} \\
 & \textbf{Urdu-Wikiann} & \textbf{Shahmukhi} & \textbf{SiNER} & \textbf{Pashto-Wikiann} \\
 \hline
\textbf{PER} & 254, 82.2\% & 482, 48.59\% & 555, 32.04\% & 3, 10.71\% \\
\textbf{LOC} & 102, 30.82\% & 140, 52.83\% & 115, 28.97\% & 5, 12.5\% \\
\textbf{ORG} & 234, 77.74\% & 66, 42.86\% & 57, 22.62\% & 8, 23.53\% \\
\textbf{Total} & 590, 62.69\% & 688, 48.75\% & 727, 30.53\% & 16, 15.68\% \\
\hline
\multicolumn{5}{c}{\textbf{Multilingual Datasets}} \\
 & \textbf{Urdu-Wikiann} & \textbf{Shahmukhi} & \textbf{SiNER} & \textbf{Pashto-Wikiann} \\
 \hline
\textbf{PER} & 255, 82.52\% & 507, 51.11\% & 559, 32.27\% & 6 21.43\% \\
\textbf{LOC} & 106, 32.02\% & 151, 56.98\% & 116, 29.22\% & 11 27.5\% \\
\textbf{ORG} & 234, 77.74\% & 69, 44.81\% & 57, 22.62\% & 9 26.47\% \\
\textbf{Total} & 595, 63.23\% & 727, 51.52\% & 732, 30.74\% & 26, 25.49\%\\
 \hline
\end{tabular}
\caption{Analysis of presence of named entities of test sets in monolingual and multilingual training sets. }
\label{tab:dataset-analysis}
\end{table*}

\section{Augmentation Analysis}
\label{sec:appendixB}
The cluster-based data augmentation has been performed to produce enhanced datasets with multiple iterations. The X\textsubscript{1} iteration shows a single pass of augmentation, X\textsubscript{2} iteration depicts two passes, and so on. In this section, we present an experimental analysis of the cluster-based augmentation with respect to different augmentation iterations.  

Table~\ref{tab:onevstwo-lowresource} presents the NER results from the fine-tuned Glot500 model with mono- and multilingual low-resource data settings. The micro F\textsubscript{1} scores are compared against one and two iterations. The Urdu-Wikiann dataset demonstrates some improvements for X\_2 in the monolingual setup using 100 and 200 samples. However, there is a decrease in the performance in multilingual experiments for all the other training sets. Similarly, Shahmukhi shows improved performance in monolingual setup and performance degradation in multilingual training. The SiNER and Pashto-Wikiann datasets also follow the similar trend for low-resource training splits.

Table~\ref{tab:onevstwo-wholemono} further shows NER results after fine-tuning on the entire datasets. In monolingual experiments, SiNER shows a subtle increase in scores with X\_2 iterations in both mono- and multilingual setups. However, all the other datasets demonstrate performance degradation with the increase of iterations of data augmentations. Based on these NER results, we presented results and comparisons against one iteration of data augmentation in the results section of the  paper.

\begin{table*}[t]
\small
\centering
\begin{tabular}{lcccccccc}
\hline
{ \textbf{Monolingual Setup}} & \multicolumn{2}{c}{{ \textbf{100}}} & \multicolumn{2}{c}{{ \textbf{200}}} & \multicolumn{2}{c}{{ \textbf{500}}} & \multicolumn{2}{c}{{ \textbf{1000}}} \\
\textbf{Datasets} & { \textbf{X\_1}} & { \textbf{X\_2}} & { \textbf{X\_1}} & { \textbf{X\_2}} & { \textbf{X\_1}} & { \textbf{X\_2}} & { \textbf{X\_1}} & { \textbf{X\_2}} \\
\hline
{ Urdu-Wikiann} & { 76.62} & { 76.48} & { 81.00} & { 82.32} & { 83.78} & { 83.13} & { 85.31} & { 84.73} \\
{ Shahmukhi} & { 60.78} & { 62.24} & { 68.03} & { 68.79} & { 73.17} & { 73.03} & { 77.11} & { 78.15} \\
{ SiNER} & { 65.64} & { 65.66} & { 71.17} & { 70.84} & { 76.90} & { 78.67} & { 79.46} & { 79.77} \\
{ Pashto-Wikiann} & { 48.54} & { 48.51} & -- & -- & -- & -- & -- & --\\ 
\hline
\textbf{Multilingual Setup} & \multicolumn{2}{c}{\textbf{100}} & \multicolumn{2}{c}{\textbf{200}} & \multicolumn{2}{c}{\textbf{500}} & \multicolumn{2}{c}{\textbf{1000}} \\
\textbf{Datasets} & \textbf{X\_1} & \textbf{X\_2} & \textbf{X\_1} & \textbf{X\_2} & \textbf{X\_1} & \textbf{X\_2} & \textbf{X\_1} & \textbf{X\_2} \\

\hline
Urdu-Wikkiann & 76.83 & 70.81 & 82.25 & 74.75 & 84.35 & 81.79 & 85.34 & 84.27 \\
Shahmukhi (1k) & 68.88 & 65.55 & 73.47 & 70.59 & 77.51 & 75.56 & 80.01 & 79.52 \\
Sindhi & 66.76 & 68.77 & 73.22 & 71.72 & 76.26 & 75.89 & 79.61 & 79.01 \\
Pashto & 66.63 & 68.67 & 73.12 & 71.60 & 76.05 & 75.68 & 79.35 & 78.81\\
\hline
\end{tabular}
\caption{Micro-F\textsubscript{1} scores by fine-tuning Glot500-base on low-resource multilingual datasets by using data augmentation with one (X\_1) and two (X\_2) iterations.}
\label{tab:onevstwo-lowresource}
\end{table*}

\begin{table*}[]
\small
\centering
\begin{tabular}{lcccccc}
\hline
\textbf{Monolingual Setup} & \multicolumn{3}{c}{\textbf{X\_1}} & \multicolumn{3}{c}{\textbf{X\_2}} \\
\textbf{Datasets} & \textbf{Precision} & \textbf{Recall} & \textbf{F\_1} & \textbf{Precision} & \textbf{Recall} & \textbf{F\_1} \\
\hline
Urdu-Wikiann & 94.51 & 94.61 & 94.56 & 93.75 & 94.14 & 93.94 \\
Shahmukhi & 84.04 & 82.23 & 83.13 & 82.33 & 82.20 & 82.27 \\
SiNER & 87.49 & 88.82 & 88.15 & 88.91 & 88.01 & 88.48 \\
Pashto-Wikiann & 52.08 & 45.45 & 48.54 & 58.46 & 41.45 & 48.51 \\
\hline
\textbf{Multilingual Setup} & \multicolumn{3}{c}{\textbf{X\_1}} & \multicolumn{3}{c}{\textbf{X\_2}} \\
\textbf{Datasets} & \textbf{Precision} & \textbf{Recall} & \textbf{F\_1} & \textbf{Precision} & \textbf{Recall} & \textbf{F\_1} \\
\hline
Urdu-Wikiann & 96.07 & 96.18 & 96.12 & 94.76 & 95.70 & 95.23 \\
Shahmukhi & 89.03 & 85.50 & 87.22 & 86.83 & 86.08 & 86.45 \\
SiNER & 89.19 & 86.69 & 87.92 & 88.16 & 88.01 & 88.08 \\
Pashto-Wikiann &89.00 & 86.29 & 87.62 & 87.85 & 87.54 & 87.69  \\
\hline
\end{tabular}
\caption{Micro-F\textsubscript{1} scores by fine-tuning Glot500-base on multilingual setting for the entire datasets by using data augmentation with one (X\_1) and two (X\_2) iterations.}
\label{tab:onevstwo-wholemono}
\end{table*}

Additionally, we compared the data augmentation method by selecting all correct sentences from the top five candidates with one and two iterations. Table~\ref{tab:onevstwovsall-lowres} shows the comparison for low-resource settings. In the low-resource datasets, Urdu-Wikiann and Shahmukhi datasets perform better for only 100 samples for both mono- and multilingual experiments. The other data splits start performance degradation. SiNER demonstrates some improvements for 1,000 sentences in monolingual experiment and for 100 train samples for multilingual setup. The performance degradation is observed for all the other training sets. Pashto-Wikian is a smaller dataset that contains only 100 sentences and it shows improvements by learning cross-lingual representations in multilingual setup.  

\begin{table*}[tp]
\small
\centering
\begin{tabular}{clcccc}
\hline
\multicolumn{1}{l}{\textbf{Train Size}} & \textbf{Iteration} & \textbf{Urdu-Wikiann} & \textbf{Shahmukhi} & \textbf{SiNER} & \textbf{Pashto-Wikiann} \\

\multicolumn{6}{c}{\textbf{Monolingual Setup}} \\
\hline
\multirow{3}{*}{\textbf{100}} & \textbf{X\_1} & 76.62 & 60.78 & 65.64 & 48.54 \\
 & \textbf{X\_2} & 76.48 & 62.25 & 65.66 & 48.51 \\
 & \textbf{All correct} & 72.31 & 64.39 & 65.27 & 49.78 \\
\hline
\multirow{3}{*}{\textbf{200}} & \textbf{X\_1} & 81.00 & 68.03 & 71.17 & -- \\
 & \textbf{X\_2} & 82.32 & 68.79 & 70.84 & -- \\
 & \textbf{All correct} & 81.18 & 67.85 & 71.46 & -- \\
 \hline
\multirow{3}{*}{\textbf{500}} & \textbf{X\_1} & 83.78 & 73.17 & 76.88 & -- \\
 & \textbf{X\_2} & 83.13 & 73.03 & 78.67 & -- \\
 & \textbf{All correct} & 84.57 & 73.87 & 76.00 & -- \\
 \hline
\multirow{3}{*}{\textbf{1000}} & \textbf{X\_1} & 85.31 & 77.11 & 79.46 & -- \\
 & \textbf{X\_2} & 84.73 & 78.15 & 79.77 & -- \\
 & \textbf{All correct} & 81.76 & 77.16 & 80.98 & -- \\
 \hline
\multicolumn{6}{c}{\textbf{Multilingual Setup}} \\
\hline
\multirow{3}{*}{\textbf{100}} & \textbf{X\_1} & 76.83 & 66.88 & 66.76 & 66.63 \\
 & \textbf{X\_2} & 70.81 & 65.55 & 68.77 & 68.67 \\
 & \textbf{All correct} & 79.10 & 67.85 & 64.89 & 64.84 \\
\hline
\multirow{3}{*}{\textbf{200}} & \textbf{X\_1} & 82.25 & 73.47 & 73.22 & 73.12 \\
 & \textbf{X\_2} & 74.75 & 70.59 & 71.72 & 71.60 \\
 & \textbf{All correct} & 79.58 & 71.55 & 72.84 & 72.70 \\
\hline
\multirow{3}{*}{\textbf{500}} & \textbf{X\_1} & 84.35 & 77.51 & 76.26 & 76.05 \\
 & \textbf{X\_2} & 81.79 & 75.56 & 75.89 & 75.68 \\
 & \textbf{All correct} & 81.49 & 76.00 & 77.03 & 76.86 \\
\hline
\multirow{3}{*}{\textbf{1000}} & \textbf{X\_1} & 85.34 & 80.01 & 79.61 & 79.35 \\
 & \textbf{X\_2} & 84.27 & 79.52 & 79.01 & 78.81 \\
 & \textbf{All correct} & 85.03 & 79.13 & 79.18 & 79.53 \\
\hline
\end{tabular}
\caption{Micro-F\textsubscript{1} scores by fine-tuning Glot500-base on monolingual and multilingual low-resource datasets by using data augmentation with one (X\_1) and two (X\_2) iterations and all correct from top five augmentations.}
\label{tab:onevstwovsall-lowres}
\end{table*}

We further compared the results by selecting all correct sentences for entire datasets as shown in Table~\ref{tab:onevstwovsall-entire}. The F\textsubscript{1} score for Urdu-Wikiann remains in the same range for monolingual training but deceases significantly in the multilingual training setup. However, F\textsubscript{1} scores for Shahmukhi and Sindhi are quite low compared to X\_1 and X\_2 iterations. Pashto-Wikiann shows the similar behaviour. 

The Shahmukhi and SiNER datasets were further analyzed for one, two and three augmentation iterations for low-resource monolingual settings as shown in Table~\ref{tab:onevstwovsthree-lowres}. Shahmukhi shows improvements by training with three iterations. However, in the multilingual setup, it shows performance degradation when adding more augmented sentences (Table~\ref{tab:onevstwovsall-entire}). On the other hand, SiNER performs with mixed results but it also demonstrates decreased performance in multilingual training setup with increased data augmentation iterations. Based on these analysis, augmentation with one iteration produces optimal performance for Urdu-Wikiann, Shahmukhi, SiNER and Pashto-Wikiann datasets. Therefore, in the main paper, we presented the results achieved by using one iteration of the cluster- and EDA-based data augmentation methods for all the selected datasets.

Table~\ref{tab:fewshot_variance} presents the F\textsubscript{1} scores for Shahmukhi and SiNER Few-Shot experiments with five different randomly selected training sets to analyze the variation in scores across datasets. Pashto-Wikiann is a small dataset with only 100 instances, and our data augmentation technique does not perform well on Urdu-Wikiann; therefore, we experimented only on the Shahmukhi and SiNER datasets. Shahmukhi exhibits a consistent trend across all Few-Shot settings, with a mean score closely aligning with the actual scores. However, SiNER, on the other hand, demonstrates higher variance for to the smaller number of examples. 

\begin{table*}[h]
\small
\centering
\begin{tabular}{lccc}
\hline
\multicolumn{4}{c}{\textbf{Monolingual Setup}} \\
\textbf{Dataset} & \textbf{X\_1} & \textbf{X\_2} & \textbf{All correct} \\
\hline
Urdu-Wikiann    & 94.56 & 93.94 & 94.58 \\
Shahmukhi   & 83.13  & 82.27 & 81.79\\
SiNER    & 88.15  & 88.48 & 86.84 \\
Pashto-Wikiann & 48.54 & 48.51 & 49.78\\
\hline
\multicolumn{4}{c}{\textbf{Multilingual Setup}} \\
\hline
Urdu-Wikiann & 96.12 & 95.23 & 91.82 \\
Shahmukhi & 87.22 & 86.45 & 83.42\\
SiNER & 87.92 & 88.08 & 84.82\\
Pashto-Wikiann & 87.62 & 87.69 & 84.52\\
\hline

\end{tabular}
\caption{Micro-F\textsubscript{1} scores by fine-tuning Glot500-base on monolingual low-resource datasets by using data augmentation with one (X\_1) and two (X\_2) iterations and all correct from top five augmentations.}
\label{tab:onevstwovsall-entire}

\end{table*}

\begin{table*}[h]
\small
\centering
\begin{tabular}{cccc}
\hline
\multicolumn{1}{l}{\textbf{Train Size}} & \textbf{Iteration} & \textbf{Shahmukhi} & \textbf{SiNER} \\
\hline
\multirow{3}{*}{\textbf{100}} & \textbf{X\_1} & 60.78 & 65.64 \\
 & \textbf{X\_2} & 62.25 & 65.66 \\
 & \textbf{X\_3} & 61.85 & 65.03 \\
 \hline
\multirow{3}{*}{\textbf{200}} & \textbf{X\_1} & 68.03 & 71.17 \\
 & \textbf{X\_2} & 68.79 & 70.84 \\
 & \textbf{X\_3} & 70.35 & 70.38 \\
 \hline
\multirow{3}{*}{\textbf{500}} & \textbf{X\_1} & 73.17 & 76.88 \\
 & \textbf{X\_2} & 73.03 & 78.67 \\
 & \textbf{X\_3} & 73.89 & 75.98 \\
 \hline
\multirow{3}{*}{\textbf{1000}} & \textbf{X\_1} & 77.11 & 79.46 \\
 & \textbf{X\_2} & 78.15 & 79.77 \\
 & \textbf{X\_3} & 77.68 & 80.53 \\
 \hline
\end{tabular}
\caption{Micro-F\textsubscript{1} scores by fine-tuning Glot500-base on monolingual low-resource datasets by using data augmentation with one (X\_1), two (X\_2) and three (X\_3) iterations.}
\label{tab:onevstwovsthree-lowres}
\end{table*}

\begin{table*}[h]
\small
\centering
\begin{tabular}{lllll}
\hline
\multicolumn{5}{c}{\textbf{Shahmukhi}} \\
\textbf{RUNs} & \textbf{100} & \textbf{200} & \textbf{500} & \textbf{1000} \\
\hline
Run 1 & 62.12 & 66.04 & 72.05 & 76.69 \\
Run 2 & 60.77 & 67.00 & 71.47 & 75.45 \\
Run 3 & 60.49 & 65.86 & 72.62 & 77.15 \\
Run 4 & 61.81 & 64.85 & 73.36 & 77.23 \\
Run 5 & 62.58 & 67.33 & 72.05 & 74.98 \\
Mean & 61.55 & 66.22 & 72.31 & 76.30 \\
Variance & 0.7963 & 0.9698 & 0.5098 & 1.0511 \\
Standard Deviation & 0.8924 & 0.9848 & 0.7140 & 1.0252 \\
\hline
\multicolumn{5}{c}{\textbf{SiNER}} \\
\textbf{RUNs} & \textbf{100} & \textbf{200} & \textbf{500} & \textbf{1000} \\
\hline
Run 1 & 64.81 & 70.40 & 75.83 & 77.78 \\
Run 2 & 60.54 & 66.62 & 75.21 & 79.32 \\
Run 3 & 63.40 & 65.89 & 74.94 & 76.81 \\
Run 4 & 63.67 & 69.44 & 74.57 & 78.74 \\
Run 5 & 64.65 & 69.86 & 73.55 & 79.45 \\
Mean & 63.41 & 68.44 & 74.82 & 78.42 \\
Variance & 2.9505 & 4.1682 & 0.7155 & 1.2437 \\
Standard Deviation & 1.7177 & 2.0416 & 0.8458 & 1.1152\\
\hline
\end{tabular}
\caption{Mean, variance, and standard deviation by fine-tuning Glot500-base for Shamukhi and SiNER Few-Shot settings on five randomly selected train sets.}
\label{tab:fewshot_variance}
\end{table*}

\section{Hyperparameters}
\label{sec:appendixD}
In the fine-tuning process, the learning rate of 2e-5 was used along with the AdamW optimizer. The batch size was set to 8, which helped to maintain memory and training efficiency. The models were fine-tuned by setting various number of epochs for low-resource datasets depending on the training samples. Early stopping was further implemented based on the micro F\textsubscript{1} score on the validation set. The maximum sequence length was set to 100 tokens. These hyperparameters ensured optimal performance of the models.

\section{Few-Shot NER - Prompt}
\label{sec:appendixE}
\textit{You are an expert in identifying named entities for {language}. The INPUT contains text followed by an OUTPUT sequence of BIO labels. Perform named entity recognition and return the labels. Three examples are provided for your reference:\\
EXAMPLE 1: \\
{INPUT}: Foreign advisor Sartaj Aziz will visit Afghanistan today.\\
{OUTPUT}: O O B-PER I-PER O O B-LOC O.}

\end{document}